\documentclass[10pt, a4paper]{article}

\usepackage[final]{lrec-coling2024} 
\usepackage{natbib}
\usepackage{makecell}
\usepackage{multibib}
\usepackage{graphicx}
\usepackage{tabularx}
\usepackage{amsmath}

\usepackage{times}
\usepackage{latexsym}
\usepackage[T1]{fontenc}
\usepackage{subcaption}
\usepackage[utf8]{inputenc}
\usepackage{graphicx} %
\usepackage{subcaption} %
\usepackage{algorithmic}
\usepackage{algorithm}
\usepackage{times}
\usepackage{latexsym}
\usepackage{tabularx}
\usepackage[T1]{fontenc}
\usepackage{booktabs}
\usepackage[utf8]{inputenc}
\usepackage{pgfplots, pgfplotstable}
\pgfplotsset{compat=1.18}
\usepackage{microtype}

\usepackage{inconsolata}

\usepackage{microtype}

\usepackage{inconsolata}

\usepackage{titlesec}
\usepackage{CJK}

\title{A Study on How Attention Scores in the BERT Model are Aware of Lexical Categories in Syntactic and Semantic Tasks on the GLUE Benchmark}

\name{Dongjun Jang, Sungjoo Byun, Hyopil Shin} 

\address{Department of Linguistics, Seoul National University \\
         3-327, 1, Gwanak-ro, Gwanak-gu, Seoul, Republic of Korea \\
         \{qwer4107, byunsj, hpshin\}@snu.ac.kr\\}

\abstract{
This study examines whether the attention scores between tokens in the BERT model significantly vary based on lexical categories during the fine-tuning process for downstream tasks. Drawing inspiration from the notion that in human language processing, syntactic and semantic information is parsed differently, we categorize tokens in sentences according to their lexical categories and focus on changes in attention scores among these categories. Our hypothesis posits that in downstream tasks that prioritize semantic information, attention scores centered on content words are enhanced, while in cases emphasizing syntactic information, attention scores centered on function words are intensified. Through experimentation conducted on six tasks from the GLUE benchmark dataset, we substantiate our hypothesis regarding the fine-tuning process. Furthermore, our additional investigations reveal the presence of BERT layers that consistently assign more bias to specific lexical categories, irrespective of the task, highlighting the existence of task-agnostic lexical category preferences.
 \\ \newline \Keywords{Linguistics, BERT, Lexical Category} }

\begin{document}

\maketitleabstract

\section{Introduction}

In the realm of sentence comprehension, human attention is not evenly distributed across all words, indicating systematic variations in language processing \citep{rayner1986lexical}. Human attention exhibits distinct and selective parsing of syntactic and semantic information, compartmentalizing of language processing into syntax and semantics \citep{dapretto1999form}.

Inspired by the intricacies of human language processing, the attention mechanism was designed to enable deep learning models to identify relevant areas of concentrated information \citep{bahdanau2014neural}. The Transformer model, which utilizes the attention mechanism, has emerged as a state-of-the-art approach \citep{vaswani2017attention}, fueling increasing interest in exploring the attention mechanism from a linguistic perspective. BERT \citep{devlin2018bert}, a prominent Transformer-based Encoder model, has been extensively studied, revealing that certain layers capture specific linguistic knowledge of syntax and semantics. However, research that specifically addresses the weight of attention scores within the token-to-token attention matrix from a lexical category perspective has been largely underexplored until now.

This study centers on the hypothesis that during the fine-tuning process of a pre-trained BERT model for specific downstream tasks, attention scores are substantially altered based on the relationship between lexical categories and the given downstream task. Lexical categories consist of content words and function words, with semantic information embedded within content words and syntactic information embedded within function words \citep{neville2mills}. Investigating the variations in attention score weights with a focus on the lexical category constitutes a meaningful exploration within the context of the training process of BERT models. This exploration seeks to ascertain whether the updates in trainable parameter values during training exhibit a correlation with the token relationship from the view of lexical category. Accordingly, we differentiate between tasks in the GLUE benchmark datasets \citep{wang2018glue} that prioritize semantic elements and those that prioritize syntactic elements. During the training process for each task, if the attention scores exhibit higher weights for specific lexical categories according to the task's objective, it would provide evidence in support of our hypothesis. 

In this paper, we introduce a novel methodology for extracting linguistic information from the token-to-token attention score matrix within BERT, designed to delve into the associations among words during the self-attention mechanism when a sentence is fed into BERT. The proposed method aims to unravel the attention distribution at each layer within a multi-layer model. Following the extraction of relationships between lexical categories from the attention score matrices, we proceed to compare and analyze the attention formed among tokens in the BERT model with the attention in the pretrained BERT model.

Our experimental results show the feasibility of interpreting attention shifts in fine-tuned BERT models, with a particular emphasis on lexical categories. We validate our initial hypothesis concerning the fine-tuning process, and our supplementary inquiries unveil a compelling discovery. Additionally, we identify the persistent inclination of BERT layers towards particular lexical categories, regardless of the specific task at hand.

\section{Related Work} 

BERT's performance across a range of downstream tasks in Natural Language Processing, including linguistic tasks, has sparked numerous investigations into the encoding and decoding of linguistic information. Despite the extensive research linking BERT and linguistics \citep{rogers2021primer}, this section focuses on studies directly relevant to our own research. \citet{jawahar2019does} discovered that BERT has the ability to capture structural language information, with lower layers capturing phrase-level information, middle layers encoding syntactic features, and top layers focusing on semantic features. \citet{htut2019attention} fine-tuned BERT model on syntax-oriented and semantics-oriented datasets, aiming to identify significant shifts through the extraction of dependency relations using attention weights. While they found that BERT's attention heads tracked individual dependency types, they concluded that this observation was not universally applicable. In contrast, \citet{kovaleva2019revealing} reported a lack of noticeable attention shifts in BERT, suggesting that the attention maps might be more influenced by the pre-training tasks rather than task-specific linguistic reasoning. Their research investigated whether BERT's fine-tuning on a specific task results in self-attention patterns emphasizing particular linguistic features. Other than these studies, our research contributes to the examination of the relationship between BERT's attention score and the lexical categories.

\section{An Extracting Algorithm for Decrypting Token Relationships within Attention Scores Mapped with Linguistic Notions}

\begin{algorithm}[!htb]
\caption{Extracting Algorithm}\label{alg:ADTRAS}
\begin{algorithmic}
\STATE \textbf{function} Alg($x$)
\IF{pair of sentences in $x$}
    \STATE $E_T \leftarrow Embedding(cls, sep, x[0], x[1])$
\ELSE
    \STATE $E_T \leftarrow Embedding(cls, sep, x[0])$
\ENDIF
    
\STATE $A \leftarrow Attention(E_T)$ 

\STATE $\bar{A} \leftarrow mean(A, axis=1)$ 

\STATE $\bar{A}_{m} \leftarrow Exclude Special Tokens(\bar{A})$
\STATE $\bar{A}_{avg} \leftarrow Average Subtoken Weights(\bar{A}_{m})$
\FOR{$l$ in $layers$}
    \FOR{$E_T$ in $\bar{A}_{avg}$}
        \STATE $max_{idx} \leftarrow argmax(E_T)$
        \IF{$max_{idx} == E_T$}
            \STATE $max_{idx} \leftarrow argmax(E_T \backslash max_{idx})$
        \ENDIF
        \STATE $lexcat \leftarrow Map Category(max_{idx})$
        \STATE $f_{l}[lexcat] \leftarrow f{_l}[lexcat] + 1$
    \ENDFOR
    \STATE $R_l \leftarrow f_{l}[lexcat] / \sum(f{_l}[lexcat])$
    \STATE \textbf{return} $R$
\ENDFOR
\STATE $\sum_{k=1}^{l}R_l$

\STATE \textbf{end function}

\end{algorithmic}
\end{algorithm}
This study aims to understand the lexical categories and the attention shifts across the multiple layers of BERT, with a particular focus on the shifts in probabilistic scores among tokens within BERT's attention score matrices. We introduce the extracting algorithm, which enables the decryption of token relationships without altering attention values during the information extraction process. This algorithm possesses the feature of not distorting or compromising the values during the process of extracting attention scores that are mapped to linguistic concepts. Also, it operates within multi-layered models, similar to BERT, and aims to elucidate the interconnections between tokens that carry significant weights in the attention scores. Its application is particularly valuable for decoding the relational structure of tokens, such as lexical categories or Part-of-Speech, which constitutes the main focus of our investigation. Additionally, this algorithm facilitates the extraction and comprehension of syntactic configurations, semantic interrelationships between words, and causal correlations.

\begin{figure*}[t] 
	\centering
	\raggedright	
	\captionsetup[subfigure]{width=.85\linewidth}
	\subfloat[Pretrained Model] 
	{	\label{fig:pitts2tokyo-512} 
		\resizebox{.45\linewidth}{!}{
			\pgfplotsset{ymin=0,ymax=1.5}
                \begin{tikzpicture}
                
                    \begin{axis}[legend pos=outer north east
                        ,width=.55\linewidth
                        ,height= 5.5cm
                        ,ymin=0,ymax=1.5,
                        ,xtick=data,
                        ,xticklabels={CoLA,MRPC,SST,QQP, MNLI, WiC}
                        ,x tick label style={rotate=90,anchor=east}
                        ]
                        \addplot [ybar ,draw = blue,
                            line width = .5mm,
                            fill = blue
                        ] coordinates 
                        {(0,1.27) (1,1.32) (2,1.13) (3,1.11) (4,1.37) (5,1.33)};
            
                        \addplot [ybar ,draw = red,
                            line width = .5mm,
                            fill = red
                        ] coordinates 
                        {(0,0.38) (1,0.21) (2,0.70) (3,0.79) (4,0.17) (5,0.19)};
                        \legend{Con., Fun.}
                    \end{axis}
                \end{tikzpicture}\hfill
		}
	}
	\captionsetup[subfigure]{width=.85\linewidth}
	\subfloat[Finetuned Model] 
	{	
		\resizebox{.45\linewidth}{!}{
			\pgfplotsset{ymin=0,ymax=1.5}
    \begin{tikzpicture}
        \begin{axis}[legend pos=outer north east
            ,width=.55\linewidth
            ,height= 5.5cm
            ,ymin=0,ymax=1.5,
            ,xtick=data,
            ,xticklabels={CoLA,MRPC,SST,QQP, MNLI, WiC}
            ,x tick label style={rotate=90,anchor=east}
            ]
            \addplot [ybar ,draw = blue,
                line width = .5mm,
                fill = blue
            ] coordinates 
            {(0,1.12) (1,1.26) (2,1.15) (3,1.15) (4,1.17) (5,1.38)};

            \addplot [ybar ,draw = red,
                line width = .5mm,
                fill = red
            ] coordinates 
            {(0,0.73) (1,0.37) (2,0.65) (3,0.70) (4,0.61) (5,0.08)};
            \legend{Con., Fun.}
        \end{axis}
    \end{tikzpicture}\hfill
		}
	}
	\caption{Changes in Attention Distribution Across Lexical Categories from Pre-trained Model to Fine-tuned Model}
	\label{finetune}
\end{figure*}
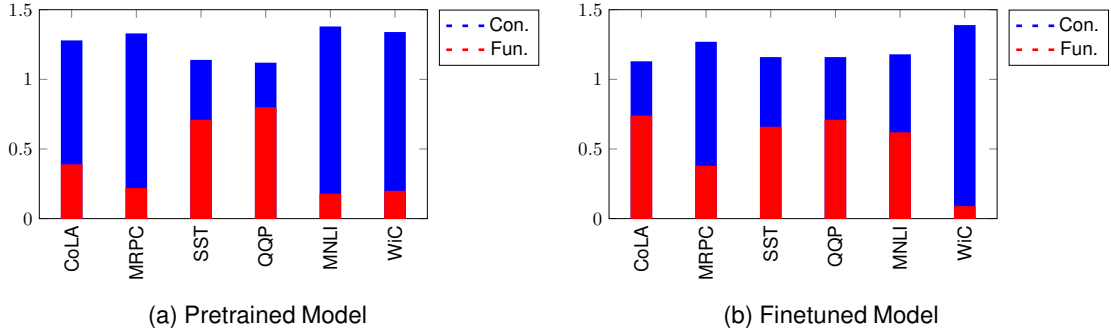

In the context of utilizing the extracting algorithm with BERT and lexical categories, the procedure begins with the application of a BERT model to tokenize and format the input sentence, denoted as $x$. This step involves incorporating specific tokens, such as \textit{CLS} and \textit{SEP}, to ensure compatibility with the BERT model. Subsequently, the algorithm obtains the self-attention weights across all layers, represented as $A$, from the BERT model and calculates the mean across the heads in each layer, denoted as $\bar{A}$. Our primary focus is on interactions involving meaningful tokens, disregarding special tokens like \textit{CLS} and \textit{SEP}. This process is referred to as

\[
{\bar{A}}_{m} = Exclude Special Tokens(\bar{A})
\]

In the event of words being segmented into sub-tokens during tokenization, the attention weights are averaged, denoted as 
\[
{\bar{A}}_{avg} = Average Subtoken Weights({\bar{A}}_{m})
\]

For each token, the algorithm locates the token $E_T$ with the most substantial attention score, $max_{score}$. In scenarios where a token's attention is chiefly self-directed, the algorithm selects the attention score ranking second highest. This selection process is denoted as $max_{idx}$ = $argmax(E_T)$ and if $max_{idx} = E_T$, then $max_{idx}$ = $argmax(E_T \setminus max_{idx})$, where $E_T \in \bar{A}_{avg}$ .

The selected tokens are subsequently categorized into their respective predetermined lexical categories, indicated as $lexcat = Map Category(max_{idx})$. The algorithm then tallies the frequency of each lexical category, represented as 
\[
f_{l}[lexcat] = f_{l}[lexcat] + 1
\]

To conclude, the relative attention ratio for each lexical category is determined by normalizing the frequency of each category by the total frequency, which helps mitigate bias. This can be mathematically represented as
\[
R_l = \frac{f_{l}[lexcat]}{\sum f_{l}[lexcat]}
\]
By deriving the attention ratios, $R$, across all layers, we can analyze the layers individually using the extracting algorithm. This allows us to perform layer-wise analysis and examine the attention distribution patterns within each layer. All these steps are summarized in Alg. \ref{alg:ADTRAS}.

\section{Experimental Setup}
For the experiment, we fine-tune the BERT-base-cased model on a selection of tasks from the GLUE benchmark \citep{wang2018glue, wang2019superglue}. We choose six diverse tasks that require different types of semantic or syntactic information. The \textbf{CoLA} task evaluates sentence grammaticality, focusing on correct syntax. \textbf{MRPC} distinguishes sentence pair equivalency, which often involves function word variation despite semantic similarity. \textbf{SST-2} detects sentiment, primarily influenced by semantic lexicons. \textbf{QQP} identifies question duplicates in pairs, typically exhibiting semantic lexicon variation. \textbf{MNLI} discerns relations (e.g., 'neutral', 'contradiction', or 'entailment') between sentences, mainly influenced by their semantic and syntactic structures. Finally, \textbf{WiC} is a 'Boolean Types' task that determines if semantic lexicon relationships categorize paired tokens as homonyms. We fine-tune the bert-base-cased model for each task and utilize the extracting algorithm to decode word attention relations, thereby highlighting notable shifts when viewed through the lens of lexical categories\footnote{We classify and tag content words and function words using the NLTK (Natural Language Toolkit) module, following the definition provided by \citet{carpenter1983eye}. See datails on Appendix \ref{sec:appendix_pos_info}}.

\section{Results}
In the results, we evaluate the six finetuned BERT models on six distinct test datasets, both before and after fine-tuning. By employing the extracting algorithm, we can discern attention variations within the lexical category at each layer.

\begin{table}
\centering
\begin{tabularx}{\linewidth}{X*{4}{c}}
\hline
\toprule
      & \multicolumn{2}{c}{\textbf{Pretrained}} & \multicolumn{2}{c}{\textbf{Finetuned}} \\
\midrule
      & Con.  & Fun.  & Con.  & Fun. \\
\midrule
\textbf{CoLA} & 1.27  & .38  & 1.12  & \textbf{.73} (\textcolor{red}{+.35})\\
\textbf{MRPC} & 1.32  & .21  & 1.26  & \textbf{.37} (\textcolor{red}{+.16})\\
\textbf{SST}  & 1.13  & .70  & \textbf{1.15}  & .65 (\textcolor{blue}{-.05})\\
\textbf{QQP}  & 1.11  & .79  & \textbf{1.15}  & .70 (\textcolor{blue}{-.09})\\
\textbf{MNLI} & 1.37  & .17  & 1.17  & \textbf{.61} (\textcolor{red}{+.44})\\
\textbf{WiC}  & 1.33  & .19  & \textbf{1.38}  & .08 (\textcolor{blue}{-.11})\\
\bottomrule
\end{tabularx}
\caption{Changes in Attention Distribution Across Lexical Categories from Pre-trained Model to Fine-tuned Model}
\label{change}
\end{table}

\subsection{Intrinsic Learning of Lexical Categories in BERT for Downstream Tasks}
This study investigates the changes in attention weights following fine-tuning for various downstream tasks, providing insights into the learning capabilities of self-attention for lexical categories. Our focus is primarily on the last layer of BERT, which is believed to be task-specific \citep{liu2019linguistic, kovaleva2019revealing, hao2019visualizing}. The results reveal significant attention shifts depending on the task type (Figure \ref{finetune} and Table \ref{change}). For example, when fine-tuning BERT for the CoLA task, which requires understanding of syntactic structures, there is an increase in attention devoted to function words, while content words experience a decrease. In contrast, fine-tuning for the WiC task, which relies on the relationships among content words, leads to an increase in attention to content words and a decrease for function words. This shift is intriguing as the model pays even more attention to content words in the fine-tuned model, despite their already significant attention in the pretrained one. Moreover, tasks like SST-2 and QQP, which prioritize semantic elements over syntactic ones, show an escalation in attention on content words. Lastly, for the MNLI task, which requires both syntactic and semantic understanding, there is a significant amplification in attention on function words. This suggests a strong association between the MNLI task and syntactic information.

In summary, we can observe a rise in attention weights for function words in tasks involving syntactic information (CoLA, MRPC, MNLI), while tasks emphasizing semantic information (SST, QQP, WiC) exhibit increased attention weights on content words in Table \ref{change}. These findings indicate that as language models undergo fine-tuning for specific objectives, they acquire intrinsic linguistic knowledge based on lexical categories.

\begin{table}[!t]
\centering
\begin{tabularx}{\linewidth}{X*{6}{c}}
\hline
    \toprule
          & \multicolumn{3}{c}{\textbf{Con.}} & \multicolumn{3}{c}{\textbf{Fun.}} \\
    \hline
        \textbf{}    & $T_1$  & $T_2$  & $T_3$  & $T_1$  & $T_2$  & $T_3$ \\
    \hline
    \midrule
        \textbf{CoLA} & L12  & L1  & L11  & L2  & L8  & L4  \\
        \textbf{MRPC} & L11  & L12  & L1 & L8  & L2  & L9 \\
        \textbf{SST}  & L1  & L11  & L12 & L8  & L2  & L4 \\
        \textbf{QQP}  & L1  & L11  & L12  & L8  & L9  & L4 \\
        \textbf{MNLI} & L12  & L11  & L1  & L8  & L2  & L4 \\
        \textbf{WiC}  & L11  & L12  & L10  & L2  & L8  & L4 \\
    \bottomrule
    \hline
\end{tabularx}
\caption{Top 3 Layers which mostly attend on the content words and function words on 6 downstream tasks}
\label{layer}
\end{table}

\subsection{Generalization of Layer-Wise Attention in Fine-Tuned BERT Models}

Table \ref{layer} offers a comprehensive breakdown of the three highest-attention layers in each fine-tuned model, emphasizing their focus on content words and function words across six distinct downstream tasks. Remarkably, despite the varying settings in which each model has been fine-tuned, the layer characteristics related to lexical categories exhibit a consistent linguistic generalization.

As depicted in Table \ref{layer}, Layers 1, 10, 11, and 12 exhibit a pronounced emphasis on content words, while Layers 2, 4, 8, and 9 demonstrate a predominant focus on function words. This discovery deviates from previous research findings that suggested BERT layers lack such generalizability \citep{htut2019attention, kovaleva2019revealing}. Leveraging our extraction algorithm, we effectively establish the linguistic attributes of BERT layers as transferable features across six diverse downstream tasks.

\section{Conclusion}
This study is grounded in the hypothesis that during the fine-tuning of a pre-trained BERT model for specific downstream tasks, attention scores experience significant shifts contingent on the relationship between lexical categories (content and function words) and the task's objectives \citep{neville2mills}. We carefully examine this phenomenon within the context of tasks sourced from the GLUE benchmark, discerning tasks that emphasize semantic elements from those focusing on syntactic one. To investigate this, we introduced a new method for extracting linguistic insights from BERT's attention score matrices. Our experimental findings robustly validate our hypothesis, offering compelling evidence of attention dynamics in fine-tuned BERT models, particularly regarding lexical categories. Furthermore, we shed light on BERT layer's innate ability to acquire linguistic knowledge associated with lexical categories during downstream tasks, underscoring the unique preferences of BERT layers for content words and function words. 

\section*{Limitations}
Despite the significant contributions of this study, it is important to acknowledge some limitations. First, our investigation focused on the BERT-based model and its attention mechanism. Second, The findings obtained through the utilization of the extracting algorithm relying on lexical categories as a proxy for capturing linguistic phenomena, which may oversimplify the intricacies of language. Furthermore, the generalizability of our findings may be limited to the specific downstream tasks and dataset used in this study. Finally, the interpretation of attention shifts and their implications may be subjective and open to different perspectives. Further research is needed to explore these limitations and expand the scope of our understanding of attention mechanisms in language processing.

\section*{Ethics Statement}
This research adheres to ethical guidelines and principles of responsible research. All experiments conducted in this study were performed in compliance with relevant regulations and guidelines, ensuring the privacy and anonymity of individuals involved. The data used in this research were obtained with proper consent and handled in accordance with ethical standards. Additionally, this study aims to contribute to scientific knowledge and understanding without causing harm or infringing upon the rights of any individuals or communities.

\nocite{*}
\section{Bibliographical References}\label{sec:reference}

\bibliographystyle{lrec-coling2024-natbib}
\bibliography{lrec-coling2024-example}

\appendix

\section{Appendix: Part-of-Speech Tagging Information}\label{sec:appendix_pos_info}

In the process of analyzing the lexical categories within our study, we rely on part-of-speech (POS) tagging to classify words into function words and content words. For the purpose of POS tagging, we utilized the Natural Language Toolkit (NLTK) module, a widely employed library in Python for natural language processing. The classification into function words and content words was based on their respective POS tags, as identified by the NLTK's POS tagger.

To ensure clarity and reproducibility of our research, we provide the complete lists of POS tags that were used to categorize words into function and content words. This categorization is pivotal for the analysis presented in our study, as it underpins the investigation of shifts in attention scores within the BERT model during the fine-tuning process. Below are the defined categories:

\subsection{Function Words}

Function words are generally characterized by their grammatical roles within sentences, contributing to the syntax rather than to the content or meaning. The following table lists the POS tags that were considered as function words in our analysis:

\begin{table}[!ht]
\centering
\begin{tabular}{ll}
\hline
\textbf{POS Tag} & \textbf{Description}                \\
\hline
CC               & Coordinating conjunction            \\
MD               & Modal                               \\
DT               & Determiner                          \\
EX               & Existential there                \\
IN               & Preposition or subordinating conjunction  \\
PDT              & Predeterminer                      \\
POS              & Possessive ending                  \\
TO               & To                                  \\
WDT              & Wh-determiner                      \\
WP               & Wh-pronoun                         \\
WP\$             & Possessive wh-pronoun              \\
WRB              & Wh-adverb                          \\
RP               & Particle                           \\
\hline
\end{tabular}
\caption{POS Tags for Function Words}
\label{tab:function_words_pos}
\end{table}

\subsection{Content Words}

Conversely, content words are known for their contribution to the meaning or content of a sentence, including nouns, verbs, adjectives, and adverbs. The table below outlines the POS tags classified as content words:

\begin{table}[t]
\centering
\begin{tabular}{ll}
\hline
\textbf{POS Tag} & \textbf{Description}                \\
\hline
NN               & Noun, singular or mass              \\
NNS              & Noun, plural                        \\
NNP              & Proper noun, singular               \\
NNPS             & Proper noun, plural                 \\
CD               & Cardinal number                     \\
FW               & Foreign word                        \\
JJ               & Adjective                           \\
JJR              & Adjective, comparative              \\
JJS              & Adjective, superlative              \\
PRP              & Personal pronoun                    \\
PRP\$            & Possessive pronoun                  \\
RB               & Adverb                              \\
RBR              & Adverb, comparative                 \\
RBS              & Adverb, superlative                 \\
VB               & Verb, base form                     \\
VBD              & Verb, past tense                    \\
VBG              & Verb, gerund/present participle     \\
VBP              & Verb, non-3rd person singular present \\
VBZ              & Verb, 3rd person singular present   \\
VBN              & Verb, past participle               \\
UH               & Interjection                        \\
\hline
\end{tabular}
\caption{POS Tags for Content Words}
\label{tab:content_words_pos}
\end{table}

The aforementioned POS tags and their classifications served as a foundational element for the lexical category analysis conducted in our study. They enabled us to insightfully mine the attention scores within the BERT model, associating them with the syntactic and semantic structures that underlie natural language.


\end{document}